# Feature-aware Conditional GAN for Category Text Generation

Xinze Li[1], Kezhi Mao[1], Fanfan Lin[2], Zijian Feng[2]

[1]*School of Electrical and Electronic Engineering, Nanyang Technological University, Singapore 639798, Singapore*

[2]*Interdisciplinary Graduate School, Nanyang Technological University, Singapore* 639798*, Singapore*

**Corresponding Author: Kezhi Mao (Email: EKZMao@ntu.edu.sg)**

**Abstract**

Category text generation receives considerable attentions since it is beneficial for various natural language processing tasks. Recently, the generative adversarial network (GAN) has attained promising performance in text generation, attributed to its adversarial training process. However, there are several issues in text GANs, including discreteness, training instability, mode collapse, lack of diversity and controllability etc. To address these issues, this paper proposes a novel GAN framework, the feature-aware conditional GAN (FA-GAN), for controllable category text generation. In FA-GAN, the generator has a sequence-to-sequence structure for improving sentence diversity, which consists of three encoders including a special feature-aware encoder and a category-aware encoder, and one relational-memory-core-based decoder with the Gumbel SoftMax activation function. The discriminator has an additional category classification head. To generate sentences with specified categories, the multi-class classification loss is supplemented in the adversarial training. Comprehensive experiments have been conducted, and the results show that FA-GAN consistently outperforms 10 state-of-the-art text generation approaches on 6 text classification datasets. The case study demonstrates that the synthetic sentences generated by FA-GAN can match the required categories and are aware of the features of conditioned sentences, with good readability, fluency, and text authenticity.

## 1. Introduction

Category text generation aims to generate a collection of controllable texts with specified categories. Category text generation can be applied to various natural language processing tasks, such as sentiment analysis (Ji & Wu, 2020; Ortigosa-Hernández et al., 2012), style transfer (Chen et al., 2021; Toshevska & Gievska, 2021), dialogue generation (Tran & Nguyen, 2019; F. Xu et al., 2022; Zeng et al., 2021), etc.

Generally, text generation can be classified into word-level augmentation and sentence-level augmentation, as shown in Fig. 1. Word-level augmentation synthesizes new sentences by substituting, swapping, dropping, inserting, and shuffling individual words following certain strategies or utilizing neural networks. For instance, easy data augmentation (EDA) techniques provided a basic set of word-level augmentation methods (Wei & Zou, 2019). The TF-IDF-based replacing approach utilized the TF-IDF score of each word for word substitution (Xie et al., 2019). By contrast, sentence-level augmentation generates novel texts based on sentence-wise information such as contextual embeddings. For example, the conditional BERT (CBERT) adopted a large BERT model for contextual augmentation (Wu et al., 2018).

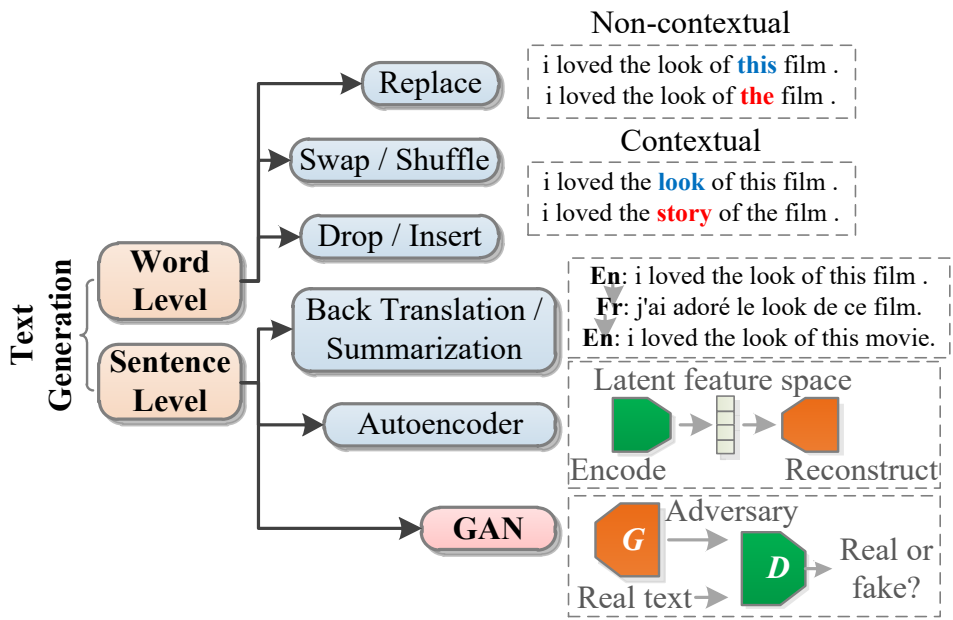

**Fig. 1.** Word-level and sentence-level text generation approaches.

One major approach for sentence-level text generation is the autoencoder, which is composed of an encoder for extracting features of conditioned sentences, and a decoder for generating synthetic texts. The encoder network embeds sentences as vectors in a latent space, and by simple vector arithmetic, novel sentences can be synthesized. The quality of sentences generated by autoencoders largely depends on the geometric property of latent space mapping. As a matter of fact, most areas in latent space are meaningless and cannot be mapped to realistic sentences (Gao et al., 2022; Mai et al., 2020). Although some improved autoencoders such as the variational autoencoder (VAE) and denoising autoencoder (DAE) have been proposed, the essential problem that the valid mapping only covers a small region of the latent space remains unresolved. In addition, most autoencoders are trained with the maximum likelihood estimation (MLE) to predict the next word, which suffers from exposure bias due to the information incompatibility between training and inference (Bengio et al., 2015). The problem of exposure bias will reduce the sentence quality, especially when the sentence is long.

The generative adversarial neural network (GAN) is a potential solution to exposure bias (Goodfellow et al., 2014). GAN introduces an adversarial min-max game between a generator and a discriminator, where the generator tries to generate realistic sentences to fool the discriminator, and conversely, the discriminator stays alert to differentiate between real and fake sentences. Since the loss function for training the generator is based on the judgments from the discriminator, the exposure bias problem is alleviated, thanks to this sequence-level adversarial strategy. Through the adversarial training, the probabilistic distribution of generated data will match the distribution of real data.

However, unlike the computer vision domain which has prosperous applications of GAN, the natural language processing (NLP) domain has only seen a few adoptions of GAN due to its three inherent characteristics. First, GAN is originally designed to generate continuous data such as images, and the natural discreteness of text induces the loss function for generators to be non-differentiable. The non-differentiability is the main reason hindering the application of GAN in NLP. To successfully apply the ideas of GAN to NLP, the challenge of discreteness should be overcome. Second, because of the additional discriminator network, the training of GAN suffers from instability. In fact, the notorious training instability will even exacerbate in discrete text domains. Moreover, the possible mode collapse issue can seriously undermine the diversity of generated sentences. How to promote the diversity of generated texts is another concern in the current literature.

To facilitate the applications of GAN to NLP, the challenges discussed above should be addressed. To handle the issue of text discreteness, reinforcement learning (RL) based approaches (Y. Li et al., 2018; K. Wang & Wan, 2018; Yu et al., 2016), and Gumbel-SoftMax based approaches (Liu et al., 2019; Nie et al., 2019) have been proposed. RL-based approaches circumvent the discreteness by enclosing the non-differentiable parts into rewards, and Gumbel-SoftMax-based approaches loosen the discreteness of the one-hot encoded output to continuous values so that the gradients from which can be directly backpropagated. In order to mitigate the training instability, different loss functions and regularization are explored. The optimization of the conventional min-max loss functions of GAN is equivalent to minimizing the Jensen-Shannon divergence between real data and synthetic data, which leads to weak learning signals when the discriminator is close to local optima (Z. Li et al., 2019). Improved adversarial training algorithms such as Wasserstein GANs (WGANs) and least squared GANs can achieve a stabler training process (Arjovsky et al., 2017). To tackle the mode collapse issue, the relational memory core (RMC) is adopted as the generator for sufficient expressiveness (Nie et al., 2019). To improve diversity of synthetic data, the discriminator to language models is modified in (J. Xu et al., 2018).

Although these GAN-based methods have shown some promising results in NLP, there is still great potential for improvements. First, most of the existing GAN-based approaches cannot controllably and reliably generate category texts. For example, the widely adopted GAN models like sequence GAN (SeqGAN) (Yu et al., 2016), relational GAN (RelGAN) (Nie et al., 2019), and diversity-promoting GAN (DP-GAN) (J. Xu et al., 2018) can only generate unlabeled texts, which may not be useful in category-related applications. Second, the structure of the generator is usually a single decoder, which samples from a prior distribution to synthesize realistic sentences. The single-decoder structure of the generator may restrict the sentence diversity. If an additional encoder is included in the generator to form a sequence-to-sequence structure, the decoder will be aware of the features of conditioned sentences, and the expressiveness of the generator can be further enhanced.

To achieve controllable category text generation for the benefits of various category-related tasks, a novel feature-aware conditional GAN (FA-GAN) is proposed in this paper. FA-GAN has several attractive characteristics that help generate diversified, novel, and realistic sentences with given categories.

First, the generator is a sequence-to-sequence network, consisting of a feature encoder, a category encoder, a word encoder, and a relational memory core (RMC) decoder with Gumbel-SoftMax output activation function. The feature encoder utilizes BERT layers to extract the contextual information of conditioned sentences. With the specially introduced feature encoder, diversified sentences can be generated, and the issue of mode collapse of GAN is mitigated. The category encoder embeds the categorical information to synthesize texts with predetermined categories. For the decoder of the generator, the RMC module instead of LSTM and GRU modules is adopted to enhance the inference capability. The Gumbel-SoftMax relaxation, instead of RL-based approaches, is adopted to solve the non-differentiable issue since RL-based approaches can cause training instability if the reward is not informative enough. For the discriminator, in addition to judging whether the input sentence is real or fake, the categories of sentences are also predicted to provide categorical information for training.

Apart from the novelty of the FA-GAN structure, the loss function in the adversarial training of FA-GAN has also been refined to cater for the categorical information of the generated sentences. The adversarial training loss includes a Wasserstein adversarial loss and a multi-class classification loss. The adopted Wasserstein adversarial loss can provide strong learning signals to stabilize the adversarial training process. With the multi-class classification loss, the generator and the discriminator will iteratively adjust the categorical information of sentences during the adversarial training process. The generator attempts to generate realistic sentences with the specified categories, while the discriminator aims to distinguish whether the sentences are real or fake and whether the categories of the sentences are the desired ones.

The organization of this paper is presented as follows. In Section 2, text generation and the applications of GAN in text domains are reviewed. Section 3 details the network structure and the adversarial training process of the proposed FA-GAN. In Section 4, the experimental results are provided, and Section 5 concludes the paper.

## 2. Related Works

The main approaches for text generation and the applications of GAN in NLP are reviewed and discussed in Section 2.1 and Section 2.2, respectively.

## 2.1 Text Generation Approaches

In general, there are two main types of text generation approaches, including word-level approaches and sentence-level approaches.

Word-level approaches adjust the individual words in a sentence to generate novel sentences, such as word-level replacement, insertion, swap, deletion, and permutation. For example, a bag of word-level text augmentation tools was developed to boost the accuracy of text classification tasks (Wei & Zou, 2019); three word-level augmentation techniques including permutation, antonym and negation were adopted to augment insufficient labeled data (Haralabopoulos et al., 2021). Uninformative words with low TF-IDF scores were substituted to achieve higher text classification accuracy (Xie et al., 2019). Sentimental words were categorized to refine the probability of words for a given emotion (Duan et al., 2020). These approaches perform word-level augmentations either randomly or by following some criteria such as TF-IDF scores, while the contextual information of each word is not considered, which can lead to harmful category shifts. To realize contextual augmentation, popular contextual embedding networks such as BERT can be adopted. For instance, a bidirectional RNN language model was used for word-level contextual augmentation (Kobayashi, 2018). BERT framework was retrofitted to the conditional BERT (CBERT), which was fine-tuned to predict masked words given both the context and the label (Wu et al., 2018). With CBERT, various text classification tasks have seen obvious improvements.

Compared with word-level augmentations, sentence-level approaches are more popular since they can maintain the global semantics of a sentence. Most of the sentence-level augmentations are sequence-to-sequence networks. For examples, back translation and abstractive summarization are widely used for novel text generation. Back translation was adopted to synthesize paraphrased sentences to boost machine translation tasks (Sennrich et al., 2015; Z. Yang et al., 2019). A back-and-forth translation approach was proposed for text data augmentation in sentiment classification tasks (Body et al., 2021). Abstractive summarization methods summarize the input texts with novel linguistic expressions, and a new attention mechanism for the abstractive summarization of social media short texts was proposed (Q. Wang & Ren, 2021).

In addition, autoencoders are dominant in the existing sentence-level text augmentation approaches, since novel sentences can be easily constructed by feature vector arithmetic. The latent feature space and the sentence generation capability of text autoencoders were studied (Bowman et al., 2015). A serial robust autoencoder was adopted for textual domain adaptation (S. Yang et al., 2020). VAE was applied in text classification and the accuracy was improved by 1-3% (Moreno-Barea et al., 2020). As a more robust approach, DAE introduces noises to inputs to preserve proper latent space geometry (Ashfahani et al., 2020; Fahfouh et al., 2020; Shen et al., 2019). Similar to DAE, a corruption function was used to perturb samples off the data manifold and a subsequent reconstruction function to project corrupted samples back,

## 2.2 Applications of GAN in Text Domains

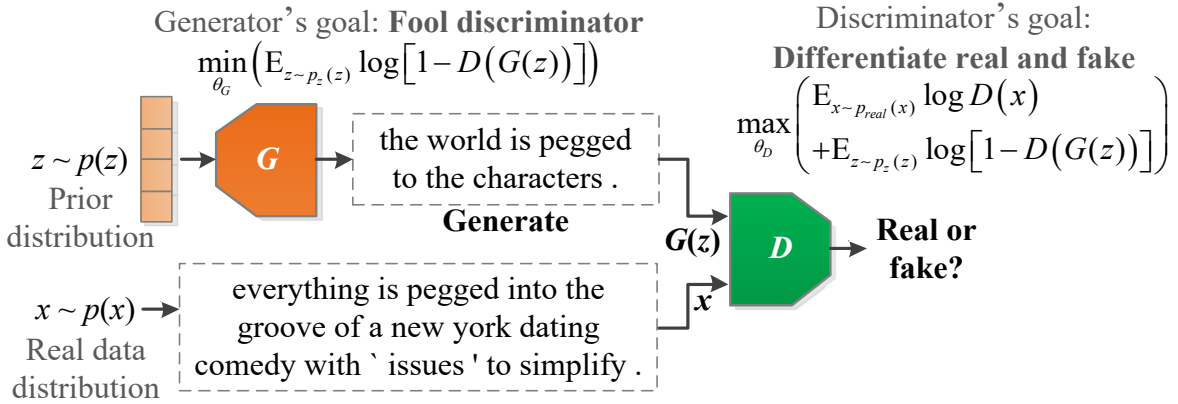

**Fig. 2.** The schematic of the conventional GAN.

Although autoencoders have been proven to be useful for general text augmentation, the underlying problem of exposure bias can severely deteriorate the quality of the generated sentence. Exposure bias is mainly caused by the incompatible information between the training process and the inference process. During the MLE training, ground-truth words are used to predict the next word. In contrast, during the inference process, the next word is predicted based on the previously generated words, and the ground-truth words are not provided. Consequently, the quality deteriorates quickly as the sentence becomes longer due to accumulated error (Bengio et al., 2015).

To alleviate the exposure bias problem, GAN is a potential solution. The conventional GAN has a generator to continuously synthesize sentences to deceive the discriminator and a discriminator to distinguish real and fake sentences, as shown in Fig. 2 (Goodfellow et al., 2014). The adversarial optimization objective of a standard GAN is given in Eqn. (1), where $G$ represents the generator network, $\theta_G$ is the parameters of $G$, $D$ is the discriminator network, $\theta_D$ is the parameters of $D$, $p_{real}(x)$ is the real data distribution, and $p(z)$ is a prior distribution. New sentences are generated by sampling the prior $p(z)$ and feeding the latent representation $z$ into the generator $G(z)$. In text domains, the standard framework of GAN suffers from the issues of non-differentiable discreteness, training instability, and mode collapse.

$$\min_{\theta_G} \max_{\theta_D} \left( \mathrm{E}_{x \sim p_{real}(x)} \log D(x) + \mathrm{E}_{z \sim p_z(z)} \log \left[ 1 - D(G(z)) \right] \right) \tag{1}$$

In general, GAN approaches in text domains can be classified into unconditional approaches and conditional approaches. Most of the GAN approaches are unconditional approaches, which can only be used for general text generation with unknown categories or labels. For instance, SeqGAN is a popular textual GAN framework (Yu et al., 2016), which retrofits the REINFORCE algorithm and directly updates the policy gradient to bypass the non-differentiability. In addition, to facilitate diversified sentence generation, the original discriminator which distinguishes fake and real sentences to language models was modified (J. Xu et al., 2018). Moreover, RelGAN adopted the RMC module as its generator to improve the text generation capability, where the Gumbel-SoftMax relaxation avoided the non-differentiable discreteness (Nie et al., 2019). Besides, to stabilize the adversarial training process, WGAN was proposed (Arjovsky et al., 2017). An improved WGAN with gradient penalty for the text augmentation of small datasets was also proposed (Moreno-Barea et al., 2020).

Compared to the omnipresent unconditional GANs, conditional GANs are rarely reported in the literature, which can be conditioned on auxiliary attributes such as class labels. If the conditional GAN takes class labels as its condition, various sentences with specified categories can be generated (Mirza & Osindero, 2014). Several cutting-edge conditional GANs for category text generation in the literature are discussed as follows. In the category sentence GAN (CS-GAN), an auxiliary classifier was used to evaluate the sentence category to lead the generation of category texts (Y. Li et al., 2018). CS-GAN adopts the same RL-based adversarial procedure as SeqGAN, which may degrade performance when the reward signal is less informative. Moreover, the RL-based training procedure suffers from training instability. Sentiment GAN (SentiGAN) (K. Wang & Wan, 2018) utilized multiple generators to synthesize samples of different labels, and a multi-class discriminator was applied to guide the category-aware training. If the number of categories increases, the trainable parameters will dramatically increase due to the multiple generators, which may exacerbate training instability. Category-aware GAN (CatGAN) utilized an efficient category-aware model for category text generation and a hierarchical evolutionary algorithm for training (Liu et al., 2019). Although CatGAN has achieved satisfactory performance, the evolutionary training procedure is time-consuming and requires sophisticated tuning.

## 3. The Proposed FA-GAN Model

In this section, the structure and adversarial training process of the proposed FA-GAN model are presented in detail. To overcome the problems in the existing text generation approaches and to generate diversified, novel, and realistic sentences with specified categories, the FA-GAN model has the following distinctive merits. First, the generator ($G$) is a sequence-to-sequence network, which is conditioned on input sentences and predetermined categories through a feature encoder (*FeatEnc*) and a category encoder (*CatEnc*), respectively. The decoder (*Dec*) of $G$ is the RMC module with the Gumbel SoftMax output function. Second, the discriminator ($D$) judges whether the input sentences are authentic or synthetic, as well as predicts the categories. Third, in the adversarial training process, the multi-class classification loss is also taken into account to guide the category text generation of the generator. Table 1 summarizes the symbols used in the FA-GAN.

**Table 1.** Symbols.

| Symbol | Interpretation | Symbol | Interpretation |
|---|---|---|---|
| $x$ | The real sentences | $\hat{x}$ | The synthesized sentences |
| $G$ | The generator network | $D$ | The discriminator network |
| *FeatEnc* | Feature encoder | *CatEnc* | Category encoder |
| *Dec* | Decoder | $s$ | Conditioned sentences |
| $o_t$ | Output logits given input $x_t$ | $\beta$ | Inverse temperature of the Gumbel SoftMax function |
| $x_{t+1}$ | The $t+1^{th}$ word of $x$ in one-hot encoding | $\hat{x}_{t+1}$ | The $t+1^{th}$ word of $\hat{x}$ |
| $c$ | Ground-truth category of $x$ | $\hat{c}$ | The required category |
| $D_r$ | Probability of the input sentence being real | $D_{cls}$ | Category probability |
| $p_{real}(x)$ | Probability distribution of real sentences $x$ | $p_z(z)$ | A prior distribution to synthesize $\hat{x}$ |
| $\lambda_{adv,d}$, $\lambda_{cls,d}$ | Weight for the adversarial loss and multi-class loss for training $D$ | $\lambda_{adv,g}$, $\lambda_{cls,g}$ | Weight for the adversarial loss and multi-class loss for training $G$ |
| $l_d$ | Total loss for training $D$, which includes $l_{adv,d}$ and $l_{cls,d}$ | $l_g$ | Total loss for training $G$, which includes $l_{adv,g}$ and $l_{cls,g}$ |

## 3.1 The Generator Structure

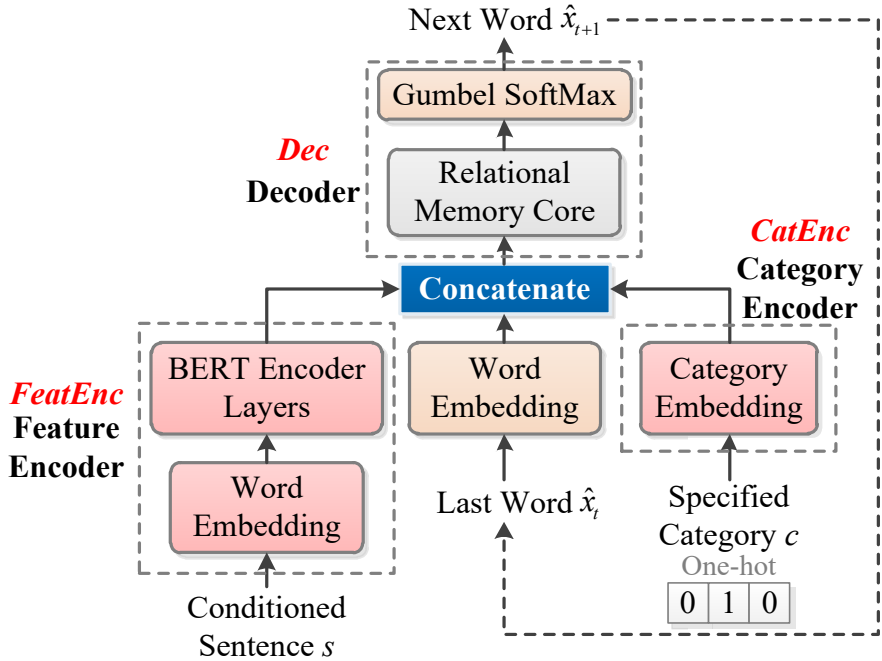

**Fig. 3.** The architecture of the generator of FA-GAN.

Fig. 3 shows the sequence-to-sequence structure of the generator in the proposed FA-GAN. The encoder of the generator is composed of three components: the feature encoder *FeatEnc*, the category encoder *CatEnc*, and the word encoder. The word encoder is an embedding layer that is intrinsically a lookup table storing the word embeddings of a fixed dictionary, which is also used in *FeatEnc* to embed individual tokens. *FeatEnc* extracts the contextual features of the conditioned sentence $s$, and *CatEnc* embeds the specified category $c$. The outputs of *FeatEnc*, the word embedding, and *CatEnc* are concatenated at the feature dimension as the feature representation. Afterwards, the decoder (*Dec*), which consists of an RMC module and a Gumbel SoftMax output activation function, iteratively gives the next words to generate a sentence with the specified category $c$.

### 3.1.1 Feature Encoder *FeatEnc*

*FeatEnc* utilizes the same word embedding layer as the word encoder and stacks several BERT encoder layers on top. Since BERT encoder layers are proficient in extracting the contextual features of the conditioned sentence (Devlin et al., 2018), they are adopted in *FeatEnc*. By stacking more BERT encoder layers, the contextual representation capability of *FeatEnc* will be stronger. Because of the specially introduced *FeatEnc*, the generator of FA-GAN can constantly refer to the contexts of the conditioned sentence to synthesize novel sentences. The generated sentences are correlated with the conditioned sentences to some extent. From the probabilistic perspective, the conventional generator in Eqn. (1) intrinsically maps a fixed latent distribution (the prior distribution) to the real data. In comparison, FA-GAN can adaptively learn the latent distribution. Consequently, benefiting from *FeatEnc*, the diversity of generated sentences is improved, and the mode collapse is relieved.

As proposed in the transformer model (Vaswani et al., 2017), the BERT encoder layer consists of a multi-head attention layer, a residual connection followed by a normalization layer, a position-wise feed-forward layer, and another residual connection followed by another normalization layer, in sequence.

### 3.1.2 Category Encoder *CatEnc*

To encode the specified category into the generator, a category encoder is adopted, which is an embedding layer as shown in Fig. 4. The one-hot encoding of the required category is viewed as the slicing index to retrieve the weights stored in the category embedding layer. The category encoder plays a significant role in guiding the generation of category texts.

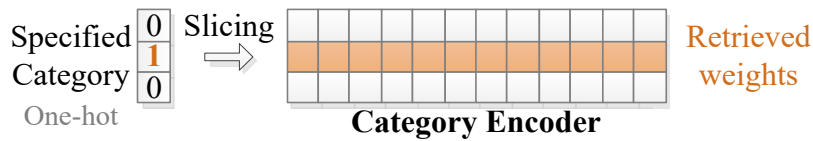

**Fig. 4.** The category encoder.

### 3.1.3 Decoder *Dec*: RMC Module with the Gumbel SoftMax Output Activation Function

In the proposed FA-GAN, the decoder generates sentences by using the RMC module with the Gumbel SoftMax output function. The RMC module instead of the LSTM module is chosen for the decoder because the RMC module is more powerful for text generation and is proven to be effective in alleviating the problems of training instability and mode collapse (Liu et al., 2019). As shown in Fig. 5, RMC considers a set of memory slots and utilizes the self-attention mechanism to interact between slots. The usage of multiple memory slots and the self-attention across slots can boost the expressiveness of the generator and handle long-distance temporal dependency (Nie et al., 2019).

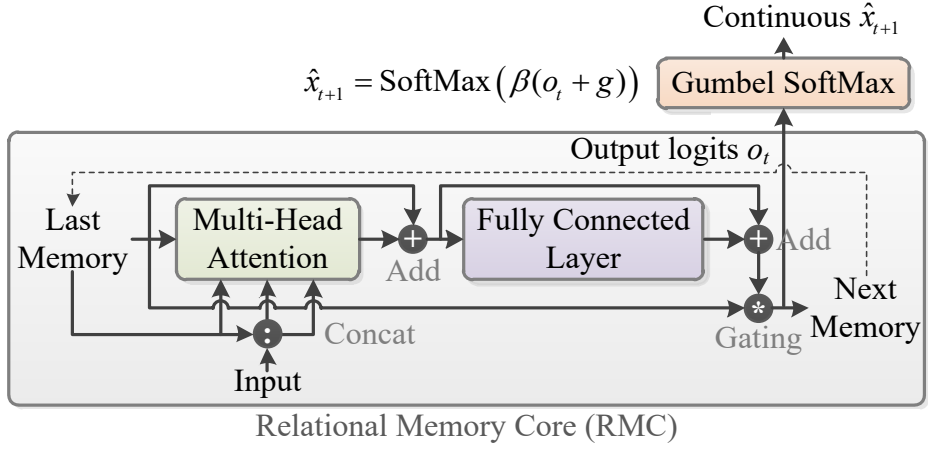

**Fig. 5.** The architecture of the decoder of FA-GAN.

To avoid the non-differentiable discreteness, the Gumbel-SoftMax relaxation is adopted in FA-GAN. RL-based approaches such as SeqGAN and SentiGAN can also tackle the issue of discreteness, but they may exacerbate the training instability since the reward from the Monte-Carlo search is sometimes uninformative. The Gumbel-SoftMax relaxation defines a continuous distribution that approximates samples from a categorical distribution as illustrated below. Conventionally, to generate the next word $x_{t+1}$ in one-hot encoding, the categorical distribution SoftMax($o_t$) in Eqn. (2) is sampled, where $o_t$ represents the output logits. However, $x_{t+1}$ is non-differentiable due to the sampling operation. The Gumbel-SoftMax relaxation to Eqn. (2) is expressed in Eqn. (3), where $\beta$ is the inverse temperature, and $g$ is a vector whose elements are independent and identically distributed random variables following the Gumbel distribution. The continuous $\hat{x}_{t+1}$ is differentiable with respect to the output logits $o_t$ and is fed into the discriminator during the adversarial training.

$$x_{t+1} \sim \text{SoftMax}(o_t) \qquad (2)$$

$$\hat{x}_{t+1} = \text{SoftMax}(\beta(o_t + g)) \qquad (3)$$

### 3.2 The Discriminator Structure

The discriminator ($D$) of FA-GAN is shown in Fig. 6, and its computational flow is discussed in this part. First, $D$ can accept real sentences $x$ and fake sentences $\hat{x}$ as inputs, which are fed into the word embedding layer. Since the fake sentence $\hat{x}$ is the output of the decoder (the RMC module with the Gumbel SoftMax function), $\hat{x}$ is continuous rather than discrete. To handle the continuous $\hat{x}$, the word embedding layer of $D$ is intrinsically a fully connected linear layer without bias. Subsequently, several BERT encoder layers extract useful information for the classification tasks. In the end, a fully connected layer with the Sigmoid activation function is responsible for differentiating whether the input sentence is authentic or fake, the output of which is denoted as $D_r$. Another fully connected layer with the SoftMax output function predicts the category probability $D_{cls}$ of the input sentence.

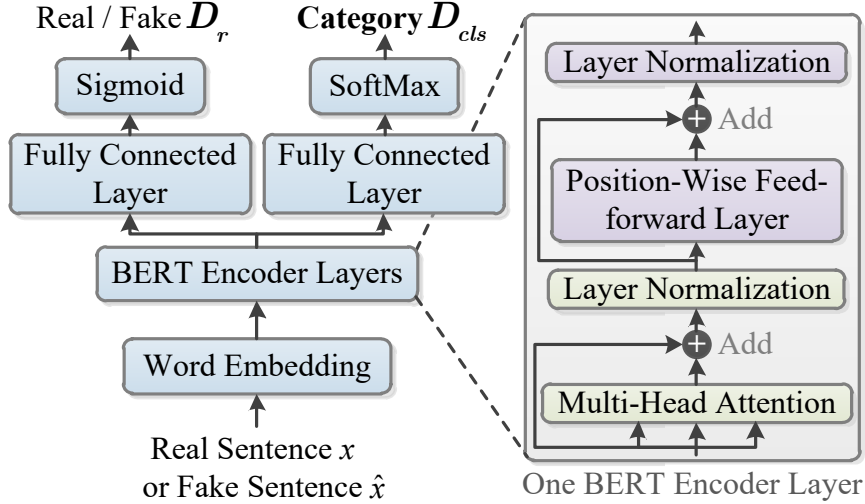

**Fig. 6.** The architecture of the discriminator of FA-GAN.

Different from the discriminator of the conventional GAN shown in Fig. 2, $D$ of FA-GAN includes two classification heads. The authenticity classification head guides the generator to synthesize realistic-looking sentences while the category classification head drives the generator to generate sentences with prespecified categories by penalizing those sentences with wrong categories. With the two classification heads, the discriminator of FA-GAN can provide effective signals regarding the sentence authenticity and category to facilitate the generator training.

### 3.3  Loss Functions in the Adversarial Training Process of FA-GAN

This section presents the loss functions in the adversarial training of FA-GAN. In general, to guide the text generation of FA-GAN towards realistic sentences with specified categories, two losses are necessary: the adversarial loss and the multi-class classification loss. The adversarial loss facilitates the authenticity of generated sentences while the multi-class classification loss encourages synthesized sentences to have predetermined categories. In FA-GAN, the Wasserstein loss is adopted as the adversarial loss to ameliorate the training instability. By contrast, the multi-class classification loss of the generated sentences is computed to train the generator $G$, and the multi-class classification loss of the real data is used for training the discriminator $D$.

### 3.3.1 Loss Functions for the Adversarial Training of the Generator

The adversarial training procedure of the generator of FA-GAN is shown in Fig. 7. First, the generator generates a batch of fake sentences $\hat{x}$ with Eqn. (4), where $z$ is sampled from a prior distribution $p_z(z)$, *FeatEnc(s)* extracts the contextual features of conditioned sentences $s$, and *CatEnc($\hat{c}$)* encodes the required categories $\hat{c}$. The generated texts $\hat{x}$ and the sampled real sentences $x$ are fed into the discriminator $D$ to compute $D_r(x)$, $D_r(\hat{x})$, and $D_{cls}(\hat{x})$, which denotes the probability of $x$ being real, the probability of $\hat{x}$ being real, and the category probabilities of $\hat{x}$, respectively. The Wasserstein adversarial loss for $G$ is given in Eqn. (5), which can mitigate the issue of weak learning signals at saddle points. Eqn. (6) computes the multi-class classification loss of the generated sentences $\hat{x}$, where $D_{cls,i}(\hat{x})$ is the probability of $\hat{x}$ belonging to the $i^{th}$ category, and $\hat{c}_i$ is the $i^{th}$ component of the one-hot encoded $\hat{c}$. The total loss for the adversarial training of $G$ is expressed in Eqn. (7), where $\lambda_{adv}$ and $\lambda_{cls}$ are adjustable weights for the adversarial loss and the multi-class loss, respectively.

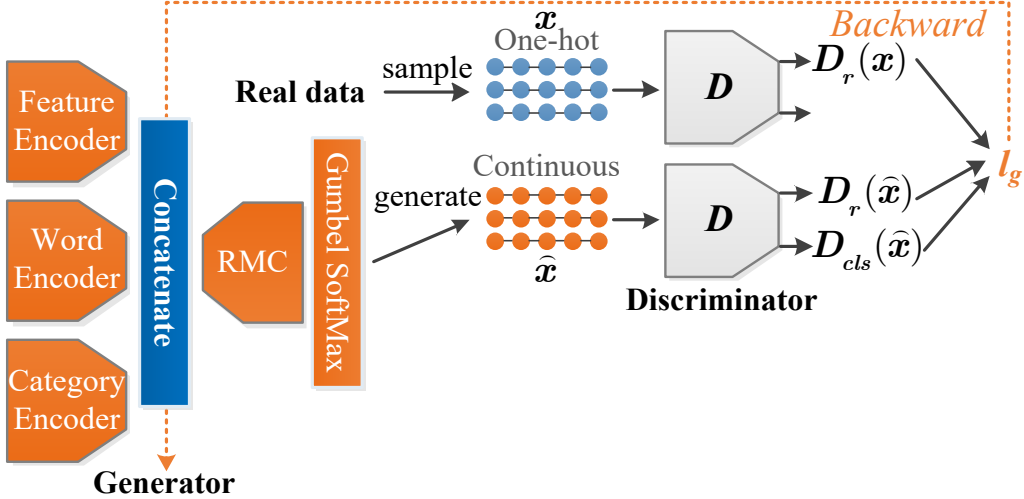

**Fig. 7.** The adversarial training of the generator.

$$\hat{x} = Dec\big(z \big| FeatEnc(s), CatEnc(\hat{c})\big), z \sim p_z(z) \tag{4}$$

$$l_{adv,g} = -\mathbb{E}_{z \sim p_z(z)}\big[D_r(\hat{x})\big] + \mathbb{E}_{x \sim p_{real}(x)}\big[D_r(x)\big] \tag{5}$$

$$l_{cls,g} = \mathbb{E}_{z \sim p_z(z)}\left[-\sum_i^k \hat{c}_i \log\big[D_{cls,i}(\hat{x})\big]\right] \tag{6}$$

$$l_g = \lambda_{adv} l_{adv,g} + \lambda_{cls} l_{cls,g} \tag{7}$$

### 3.3.2 Loss Functions for the Adversarial Training of the Discriminator

$$l_{adv,d} = \mathbb{E}_{z \sim p_z(z)}\big[D_r(\hat{x})\big] - \mathbb{E}_{x \sim p_{real}(x)}\big[D_r(x)\big] \tag{8}$$

$$l_{cls,d} = \mathbb{E}_{x \sim p_{real}(x)} \left[ -\sum_{i}^{k} c_i \log \left[ D_{cls,i}(x) \right] \right] \quad (9)$$

$$l_d = \lambda_{adv} l_{adv,d} + \lambda_{cls} l_{cls,d} \quad (10)$$

The adversarial training procedure of the discriminator of FA-GAN is shown in Fig. 8. To begin with, a batch of synthetic sentences $\hat{x}$ is generated with Eqn. (4). Afterwards, the discriminator $D$ takes $\hat{x}$ and $x$ as inputs to calculate $D_r(x)$, $D_r(\hat{x})$, and $D_{cls}(x)$, where $D_{cls}(x)$ represents the category probabilities of $x$. The Wasserstein adversarial loss for $D$ is shown in Eqn. (8), and the multi-class classification loss of the real sentences $x$ is computed with Eqn. (9), where $D_{cls,i}(x)$ is the probability of $x$ belonging to the $i^{th}$ category, $c$ is the ground-truth category of $x$ in one-hot encoding, and $c_i$ is the $i^{th}$ component of $c$. The total loss for the adversarial training of $D$ is expressed in Eqn. (10).

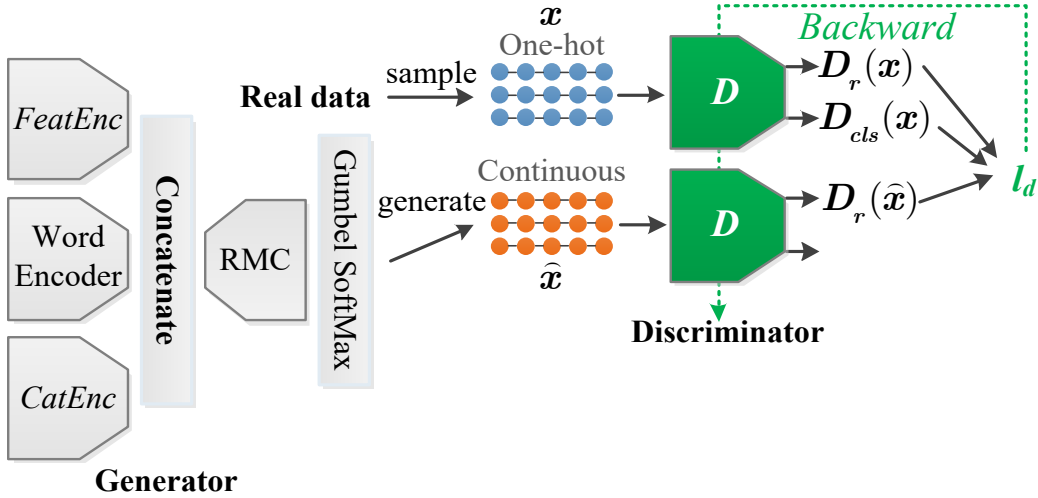

**Fig. 8**. The adversarial training of the discriminator.

The complete training procedure of FA-GAN is given in Algorithm 1. Before the adversarial training, $G$ is pre-trained to predict the next word using the MLE criterion, and $D$ is pre-trained to predict the categories of real sentences and to differentiate between authentic and synthetic sentences. During the adversarial training, $G$ and $D$ are updated by minimizing Eqns. (7) and (10), respectively. The iterative adversarial training process stops when FA-GAN reaches convergence.

### 3.4 Key Merits of FA-GAN: Category-aware and Feature-aware

In a nutshell, the proposed FA-GAN has two key merits that are distinctive from the existing GAN approaches for text generation. First, FA-GAN is category-aware, which can reliably achieve controllable category text generation. The category encoder *CatEnc* embeds the specified categories to guide the generation towards category texts. Second, FA-GAN is feature-aware, which can generate new sentences based on the contextual features of the conditioned sentences. The feature encoder *FeatEnc* greatly enhances the expressiveness of the generator and improves the diversity of generated sentences. From the perspective of loss functions, the multi-class classification loss facilitates the category text generation of FA-GAN in the adversarial training.

---

**Algorithm 1**: The adversarial training of FA-GAN

**Input**: Categorical Dataset, Generator $G$, Discriminator $D$

**Output**: Well Trained $G$ and $D$

1   Initialize $G$, $D$ with random small weights
2   Pre-train $G$ to predict the next word with MLE
3   Pre-train $D$ to classify categorical data and distinguish real and fake texts
4   **repeat**
5      **for** g_steps **do**
6         Sample real sentences as conditioned sentences $s$ ;
7         Specify the required categories of generated sentences $\hat{c}$ ;
8         Generate sentences $\hat{x}$ from $G$ by Eq. (4) and sample real sentences $x$ ;
9         Feed $x$ and $\hat{x}$ into $D$ to compute $D_r(x), D_r(\hat{x}), D_{cls}(\hat{x})$ ;
10       Update $G$ by minimizing Eq. (7) ;
11     **end for**
12     **for** d_steps **do**
13       Sample real sentences as conditioned sentences $s$ ;
14       Specify the required categories of generated sentences $\hat{c}$ ;
15       Generate sentences $\hat{x}$ from $G$ by Eq. (4) and sample real sentences $x$ ;
16       Feed $x$ and $\hat{x}$ into $D$ to compute $D_r(x), D_r(\hat{x}), D_{cls}(x)$ ;
17       Update $D$ by minimizing Eq. (10) ;
18     **end for**
19  **until** FA-GAN converges
20  **return**

---

## 4. Experiments

In this section, the performance of the proposed FA-GAN is empirically validated. To evaluate the proposed FA-GAN for category text generation, several text classification tasks are adopted, where the classification accuracy is the main evaluation criterion. In the comparisons between FA-GAN and other baseline methods, FA-GAN manifests the highest expected accuracy on all datasets.

### 4.1 Experimental Datasets

In the experiment, six popular text classification datasets are adopted:
1. Movie reviews with the maximum sequence length of 10 (MR-10): MR-10 contains two categories, which are the positive and negative reviews. There are 855 positive reviews and 781 negative reviews in the training dataset, the validation dataset has 100 sentences for both categories, and the test dataset contains 339 positive texts and 317 negative texts. MR-10 contains 3915 unique words.

2. Movie reviews with the maximum sequence length of 20 (MR-20): MR-20 has 8537 unique words with the maximum sequence length of 20. The training dataset contains 1000 positive reviews and 1000 negative reviews, and the validation set includes 100 sentences for both categories. In the test dataset, there are 1072 positive reviews and 977 negative sentences.
3. Amazon reviews of automotive products and musical instruments with the maximum sequence length of 30 (AM-30): AM-30 contains 5480 unique words in total with the maximum sequence length of 30, and the two labels of AM-30 are automotive product reviews and musical instrument reviews. The training dataset has 934 automotive product comments and 832 musical instrument reviews. The validation set has 100 reviews for both labels. In the test dataset, there are 467 and 416 reviews for automotive products and musical instruments, respectively.
4. US airline reviews with the maximum sequence length of 20 (USAir-20): There are 4991 unique words in USAir-20, and the two categories are negative reviews and positive reviews for US airlines. The training and test datasets have 800 reviews for both categories.
5. Sentiment 140 dataset with the maximum sequence length of 20 (Senti140-20): Senti140-20 contains positive and negative tweets about various brands and products. There are 6866 unique words in Senti140-20, and the training and test datasets have 1000 reviews for both labels.
6. MR-10-Low dataset tests the performance of FA-GAN under low-data regime. MR-10-Low consists of 100 sentences for each category sampled from the original MR-10 dataset.

The six datasets are summarized in Table 2, where "$Cat0$" and "$Cat1$" denote negative and positive reviews, respectively, for datasets except for AM-30. For AM-30 datasets, the two categories represent automotive product reviews and musical instrument reviews, respectively.

**Table 2.** Summarization of the datasets.

| Dataset | Unique Words | Training Size | | Validation Size | | Test Size | |
|---|---|---|---|---|---|---|---|
| | | $Cat0$ | $Cat1$ | $Cat0$ | $Cat1$ | $Cat0$ | $Cat1$ |
| MR-10 | 3915 | 781 | 855 | 100 | 100 | 317 | 339 |
| MR-20 | 8537 | 1000 | 1000 | 100 | 100 | 977 | 1072 |
| AM-30 | 5480 | 934 | 832 | 100 | 100 | 467 | 416 |
| USAir-20 | 4991 | 800 | 800 | 100 | 100 | 800 | 800 |
| Senti140-20 | 6866 | 1000 | 1000 | 100 | 100 | 1000 | 1000 |
| MR-10-Low | 2015 | 100 | 100 | 100 | 100 | 100 | 100 |

The six datasets are used to validate different properties of the proposed FA-GAN model. The MR-10 and MR-10-Low datasets aim to validate the performance of FA-GAN on short sentences, while the latter focuses on low-data regime. The MR-20, USAir-20, AM-30, and Senti140-20 datasets are used to validate the readability, fluency, and authenticity of the synthesized texts from FA-GAN considering longer temporal dependency. Besides, compared to other datasets whose categories are emotional reviews, the AM-30 dataset partitions the data into automotive products and musical instruments, which facilitate validation of the data augmentation capability of FA-GAN on general text classification task. All the datasets comprehensively validate the various properties of the proposed FA-GAN.

## 4.2 Baseline Methods

The proposed FA-GAN model is compared with 10 baseline models as summarized below.

1. EDA[1] (Wei & Zou, 2019): EDA is a generalized set of text augmentation methods, including synonym replacement, random insertion, random swap, and random deletion. EDA is easy to implement and achieves substantial improvements in low-data situations.

2. TF-IDF[2] replacement (Xie et al., 2019): TF-IDF replacement retains keywords and substitutes uninformative words with other uninformative words based on the TF-IDF criterion. Since only uninformative words are replaced, this approach aims to generate both diverse and valid samples.

3. Contextual replacement with BERT[3] (Kobayashi, 2018): To maintain the global semantics, contextual information is considered to replace words. This method feeds surrounding words to the BERT language model to find the most suitable replacement for augmentation.

4. Back and forth translation[3] (Body et al., 2021): This approach leverages several translation models for text augmentation. The key idea is to translate the original sentence in English to other intermediate languages and then translate it back.

5. SSMBA[4] (Ng et al., 2020): SSMBA is a semi-supervised text augmentation approach that perturbs the training examples off the data manifold and reconstructs the noisy point back using the pre-trained BERT model. It requires no task-specific knowledge or fine-tuning.

6. CBERT[5] (Wu et al., 2018): CBERT is a conditional BERT model for controllable category text generation. The pre-trained BERT model is fine-tuned on masked words, and the fine-tuned model can provide various words on masked places to synthesize novel sentences.

7. T5[6] (Raffel et al., 2019): T5 leverages the power of transfer learning for NLP by introducing a unified text-to-text framework in handling text-based language problems. Considering the background of text augmentation, the T5 model is applied to perform back and forth translation.

---

[1] https://github.com/jasonwei20/eda_nlp
[2] https://github.com/google-research/uda
[3] https://github.com/makcedward/nlpaug
[4] https://github.com/nng555/ssmba
[5] https://github.com/1024er/cbert_aug
[6] https://huggingface.co/docs/transformers/model_doc/t5

8. GPT-2[7] (Radford et al., 2019): GPT-2 is a large language model trained on millions of webpages in unsupervised manners. The pre-trained GPT-2 is fine-tuned on downstream tasks for topic specific text generation, which can be regarded as a reliable source of text augmentation.

9. SentiGAN[8] (K. Wang & Wan, 2018): SentiGAN is a popular GAN framework for category text generation, which consists of multiple generators and a multi-class discriminator. Each generator is responsible for a category, and a penalty-based classification loss guides the adversarial training process.

10. CatGAN[8] (Liu et al., 2019): CatGAN is an efficient category-aware GAN framework, which directly reduces the gap between real data and synthesized texts in each category to generate high-quality category sentences. A hierarchical evolutionary learning algorithm is used to train the CatGAN.

**4.3 Model Configurations**

In the experiments, a common classification model based on BERT encoders is built and adopted for all the considered approaches for fair comparisons. The classification model consists of an embedding layer, several layers of BERT encoders, a max-pooling layer, and a fully connected layer. The embedding dimension is 64, and there are 2 layers of BERT encoders whose multi-head attention layer has 2 heads with the hidden size of 64, and the intermediate hidden size of the position-wise feed-forward layer is 256. The training of the BERT-based classification model utilizes the cross-entropy loss and adopts the Adam optimizer with the learning rate of 0.0001.

The configurations of the compared baselines are as follows. In EDA, the probabilities of synonym replacement, word swap, word deletion, and word insertion are 30%, 15%, 15%, and 0% respectively; and for each real sentence, 32 synthetic sentences are generated. TF-IDF replacement approach replaces 30% of words and retains the top 5 tokens to randomly generate 8 synthetic sentences. In the contextual replacement with BERT, 8 synthetic samples are generated per real sentence, and the probability of individual words being contextually substituted is 30%. In the back and forth translation, two intermediate languages are considered: Deutsch (DE) and Russian (RU). The configurations of CBERT and SSMBA approaches are set to the default values recommended in their original papers (Ng et al., 2020; Wu et al., 2018). Besides, the text-to-text model T5 converts the original English sentence to Deutsch and French and then translates it back to English for text augmentation. In addition, to generate coherent sentences tailored for the specific task, the large language model GPT-2 is carefully fine-tuned on the datasets, and synthetic texts are sampled for augmenting the text classification accuracy. As for GAN approaches, with the settings given in (K. Wang & Wan, 2018) and (Liu et al., 2019), SentiGAN and CatGAN are carefully trained to generate realistic-looking category sentences to boost the text classification accuracy. To avoid experimental occasionality, all the baseline models have been run 10 times to evaluate the classification accuracy on test datasets.

---

[7] https://github.com/prakhar21/TextAugmentation-GPT2
[8] https://github.com/williamSYSU/TextGAN-PyTorch

In terms of the configuration of the proposed FA-GAN, the word embedding dimension is 64, and the category embedding size is 32. The structures of the *FeatEnc* in *G* and the encoder in *D* are the same, consisting of 2 BERT encoder layers. A BERT encoder layer in FA-GAN includes a two-headed attention layer with the hidden size of 64, and a position-wise feed-forward layer with the intermediate hidden dimension of 256. The RMC module in the decoder has 2 memory slots and 2 attention heads with the head size of 256. The weights of adversarial and multi-class losses ($\lambda_{adv}$ and $\lambda_{cls}$) are both 1. FA-GAN adopts the Adam optimizer with the learning rate of 0.0001 for the pre-training and the adversarial training. To enhance the adversarial training stability, the real and fake labels for *D* are occasionally flipped when training *D* (Salimans et al., 2016).

## 4.4 Experimental Results

### 4.4.1 Classification Accuracy

For fair comparison, all the compared methods are run 10 times, and the average results are reported in Table 3. In addition, t-Tests with the significance level of 0.05 are conducted to statistically validate the performance improvement of the proposed FA-GAN. Generally, on MR-10 and AM-30, FA-GAN is better than all other approaches, and it improves the classification accuracy by 1-2% on average. On MR-20, FA-GAN achieved the highest classification accuracy of 69.74%, and its performance is statistically the same as the cutting-edge CBERT model and the large language model GPT-2. On the Senti140-20 dataset, the proposed FA-GAN is better than the back-and-forth translation, SSMBA, T5, SentiGAN, and CatGAN by more than 1.3%, and it also obtains higher accuracy than other approaches. On USAir-20, FA-GAN ranks second, and the t-Test has proven that it achieves the same performance as the best SSMBA approach. Under the low-data regime, the proposed FA-GAN boosts the accuracy of the original non-augmented approach by 2.58% on MR-10-Low, and is significantly better than other approaches except for SentiGAN. Compared to the state-of-the-art models T5 and GPT-2, FA-GAN achieves an improvement of 1%. The overall performance of the proposed FA-GAN compared with other 11 methods are also summarized in Table 3, where "Better", "Same", and "Worse" denote the number of baseline models that FA-GAN outperforms, is on-par, and underperforms, respectively.

**Table 3.** Classification accuracy on test datasets. Boldface fonts indicate the best performance. Values with underlines represent the second-best performance. "*" means that the proposed FA-GAN is statistically better than the compared baseline through t-Tests, where the degree of freedom is 9 and the significance level is 0.05.

| Model | MR-10 | MR-20 | AM-30 | USAir-20 | Senti140-20 | MR-10-Low |
|---|---|---|---|---|---|---|
| No Augmentation | 72.84%* | 67.76%* | 85.53%* | 85.65%* | 71.34%* | 59.74%* |
| EDA | 72.69%* | 68.05%* | 86.50%* | 86.01%* | 72.53%* | 61.82%* |
| TF-IDF Replacement | 72.90%* | 67.91%* | 85.79%* | 86.45%* | 72.34%* | 60.34%* |
| Contextual Replacement (BERT) | 73.33%* | 68.69%* | 85.88%* | 85.81%* | 72.28%* | 60.67%* |
| Back and Forth Translation | 72.94%* | 67.85%* | 85.96%* | 86.38%* | 71.31%* | 61.51%* |
| SSMBA | 73.00%* | 68.61%* | 85.72%* | **88.51%** | 71.37%* | 59.83%* |
| CBERT | 73.29%* | 69.57% | 86.68%* | 86.20%* | 72.67% | 60.95%* |
| T5 | 73.05%* | 68.82%* | 86.72%* | 86.62%* | 71.60%* | 61.24%* |
| GPT-2 | 73.21%* | 69.69% | 86.98%* | 87.29%* | 72.61% | 61.22%* |
| SentiGAN | 73.00%* | 68.32%* | 86.42%* | 88.08% | 70.97%* | 62.14% |
| CatGAN | 73.42%* | 68.14%* | 86.27%* | 88.10% | 71.56%* | 61.90%* |
| **FA-GAN** | **74.59%** | **69.74%** | **87.52%** | 88.32% | **72.97%** | **62.32%** |
| **Merits of FA-GAN** | MR-10 | MR-20 | AM-30 | USAir-20 | Senti140-20 | MR-10-Low |
| *Better* | 11 | 9 | 11 | 8 | 9 | 10 |
| *Same* | 0 | 2 | 0 | 3 | 2 | 1 |
| *Worse* | 0 | 0 | 0 | 0 | 0 | 0 |

Through the experiments on various text classification tasks, the effectiveness of the proposed FA-GAN model is empirically validated. In these experiments, FA-GAN has boosted the expected classification accuracy by 1-3%, which also verifies that FA-GAN can generate coherent, realistic, and diversified category sentences.

### 4.4.2 BLEU Scores and $NLL_{div}$

Except for the accuracy metric on text classification tasks, other performance metrics are also used to measure the quality of generated sentences of GAN-based approaches. To evaluate the authenticity of generated sentences, the bilingual evaluation understudy (BLEU) scores of 2-gram (BLEU-2) and 3-gram (BLEU-3) are applied. In addition, to evaluate the diversity of synthesized sentences, following the recommendation of (Liu et al., 2019), the negative log-likelihood of synthesized texts on the generator $NLL_{div}$ is calculated by Eqn. (11), where $p_G$ is the generated sentence distribution defined by the generator, $\hat{x}_{1:T}$ is the generated sentence, and $\hat{x}_T$ is the $T^{th}$ word of $\hat{x}_{1:T}$. $NLL_{div}$ captures the repeatability of generated samples, and the higher the $NLL_{div}$ is, the more diversified the generated sentences are.

$$\boldsymbol{NLL}_{div} = -\mathbb{E}_{\hat{x}_{1:T} \sim p_G} \left[ \log \boldsymbol{p}_G(\hat{x}_1, ..., \hat{x}_T) \right] \quad (11)$$

The BLEU-2, BLEU-3, and $NLL_{div}$ scores of GAN approaches on the six datasets are evaluated and listed in Table 4. In terms of the BLEU scores, FA-GAN obtains the highest BLEU-2 scores on all the datasets and achieves the best BLEU-3 scores on most of the datasets. Compared to the two latest GAN-based approaches, the better BLEU scores of FA-GAN indicate that the proposed FA-GAN can generate the most realistic-looking sentences. The sequence-to-sequence structure of FA-GAN enhances the expressiveness of the generator.

In terms of the $NLL_{div}$ metric, the proposed FA-GAN is significantly better than the state-of-the-art GAN approaches on all the datasets. On MR-10, FA-GAN improves the $NLL_{div}$ by more than 1. On MR-20, the $NLL_{div}$ of FA-GAN is 0.67 higher than the second-best CatGAN. Moreover, FA-GAN reaches the highest $NLL_{div}$ scores of 2.297 and 2.618 on AM-30 and USAir-20, respectively. On Senti140-20, the $NLL_{div}$ of the cutting-edge SentiGAN is 0.964 lower than the proposed FA-GAN. On the low-data MR-10-Low, the $NLL_{div}$ of FA-GAN is 0.514 and 2.075 higher than SentiGAN and CatGAN, respectively. The consistently best $NLL_{div}$ scores of FA-GAN show the good diversity of generated sentences, attributed to the feature-aware encoder $FeatEnc$ of FA-GAN.

**Table 4.** The BLEU scores and $NLL_{div}$ of GAN approaches.

| Dataset | Model | BLEU-2 | BLEU-3 | $NLL_{div}$ |
|---|---|---|---|---|
| MR-10 | SentiGAN | 0.380 | 0.233 | 0.648 |
| | CatGAN | <u>0.433</u> | <u>0.250</u> | <u>0.728</u> |
| | FA-GAN | **0.560** | **0.273** | **1.818** |
| MR-20 | SentiGAN | <u>0.662</u> | 0.370 | 0.794 |
| | CatGAN | 0.657 | <u>0.382</u> | <u>0.934</u> |
| | FA-GAN | **0.674** | **0.391** | **1.604** |
| AM-30 | SentiGAN | 0.729 | 0.479 | <u>1.453</u> |
| | CatGAN | <u>0.736</u> | <u>0.480</u> | 1.410 |
| | FA-GAN | **0.748** | **0.489** | **2.297** |
| USAir-20 | SentiGAN | 0.704 | 0.443 | <u>1.399</u> |
| | CatGAN | <u>0.738</u> | **0.476** | 1.387 |
| | FA-GAN | **0.767** | <u>0.451</u> | **2.618** |
| Senti140-20 | SentiGAN | <u>0.641</u> | <u>0.352</u> | <u>1.186</u> |
|  | CatGAN | 0.545 | 0.254 | 1.112 |
|  | FA-GAN | **0.662** | **0.359** | **2.150** |
| MR-10-Low | SentiGAN | 0.241 | 0.138 | <u>2.062</u> |
|  | CatGAN | <u>0.315</u> | <u>0.154</u> | 0.501 |
|  | FA-GAN | **0.346** | **0.158** | **2.576** |

To summarize, the superb BLEU and $NLL_{div}$ scores on all the datasets comprehensively validate that the proposed FA-GAN can generate high-quality, realistic-looking, and highly diversified sentences.

### 4.5 Case Study

To illustrate that the synthesized sentences can match the required categories and are aware of the features of conditioned sentences, some exemplar sentences generated by the proposed FA-GAN on MR-20 are provided in Table 5.

Examples 1 to 3 aim to generate negative movie reviews. Generally, the generated sentences in examples 1 to 3 can accurately match the required negative categories. Specifically, in example 1, the context "newsreels and the like" in the conditioned sentence $s$ is captured and maintained in the generated sentence $\hat{x}$, and similar relative clauses are applied in both $s$ and $\hat{x}$. In the second example, the adversative relationship in $s$ ("not … dull" and "merely lacks") is transformed to "but" in $\hat{x}$. In addition, the sentimental "not … dull" is changed to "ludicrous" in $\hat{x}$, and the contextual information "lacks everything" is reserved. In example 3 for style transfer, the slightly positive sentiment "harmless diversion" in $s$ is inverted to the strongly negative sentiment "unlikable diversion" in $\hat{x}$, and the parallel structure (A, B, and C) is largely retained.

Examples 4, 5, and 6 showcase generating of positive movie reviews, and the generated sentences are indeed positive comments. In particular, in example 4, the generated sentence $\hat{x}$ keeps the positive semantics "refreshingly novel" of the conditioned sentence $s$. In example 5, the conditioned sentence $s$ discusses "life", and thus FA-GAN adopts the attributive clause "that is a distinct one laughing at life" to generate a positive comment surrounding "life". In example 6 which transfers the negative category to the positive category, the negative semantics "poor casting" is converted to the highly positive phrases "moving melodrama" and "delivers the positives".

The samples given in Table 5 show that the sentences generated by FA-GAN are aware of the contextual information of conditioned sentences, and the controllable category text generation is attained. The good readability and fluency of generated sentences from FA-GAN strongly validate the text authenticity.

For comparison, some generated sentences of the benchmark models on the MR-20 dataset are given in Table 6. Overall, the readability and diversity of the synthetic sentences from the benchmark models are unsatisfactory and are inferior to the proposed FA-GAN model. In particular, word-level text augmentation approaches such as EDA, contextual replacement, and CBERT lack diversity in the generated sentences. Although the TF-IDF replacement improves the text variety, but the readability and fluency are unsatisfactory. Sentence-level augmentation methods such as the back-and-forth translation and T5 produce coherent sentences but compromise the text diversity. SSMBA and SentiGAN synthesize almost meaningless and uninformative sentences. Even though the cutting-edge CatGAN and GPT-2 models yield good sentences, the empirical results in Table 3 have justified the merits of the proposed FA-GAN model.

**Table 5.** Generated sentences of FA-GAN on MR-20. The captured contextual information in the conditioned sentence $s$ is marked as blue, and the transferred contexts in the generated sentence $\hat{x}$ are highlighted as red. Texts with underscores mark the structural information being preserved.

| Example Index | Conditioned Sentence $s$ | Category of $s$ | Generated Sentence $\hat{x}$ | Category of $\hat{x}$ |
|---|---|---|---|---|
| 1 | decasia is what has happened already to so many silent movies , newsreels and the like . | Negative | the film that starts the misfortune is not a technological newsreels and quite like . | Negative |
| 2 | not an objectionable or dull film ; it merely lacks everything except good intentions . | Negative | this movie starts to be truly ludicrous , but it lacks everything . | Negative |
| 3 | ... routine , harmless diversion and little else . | Positive | unexpected trappings , simplistic execution ... unlikable diversion . | Negative |
| 4 | much of the way , though , this is a refreshingly novel ride . | Positive | the picture is original , and the perfectly ark is refreshingly novel . | Positive |
| 5 | a hypnotic portrait of this sad , compulsive life . | Positive | a good film that is a distinct one laughing at life . | Positive |
| 6 | an ambitious , guilt-suffused melodrama crippled by poor casting . | Negative | a moving melodrama , clooney 's just delivers the positives . | Positive |

**Table 6.** Generated sentences of the compared models on MR-20.

| Model | Generated Sentence | Model | Generated Sentence |
|---|---|---|---|
| EDA | decasia is what has take place already to so many still movie newsreel and the similar . | TF-IDF Replacement | decasia is shadows has ferrara already fizzles twinkie many radical movies , newsreels and the claims . |
| Contextual Replacement | decasia is what has turned out on this many silent albums, newsreels and similar like . | Back and Forth Translation | decasia is what has happened to so many silent movies , newsreels and the like . |
| **SSMBA** | storyboard advisers has apparently already featured into by televisions , newsreels and online blogs . | **CBERT** | decasia is what has happened already to so many quiet movies , newsreel and the like . |
| **T5** | decasia is what has happened already to so many silent movies , newsreels and the like . | **GPT-2** | decasia is not likely to appeal to a large audience . |
| **SentiGAN** | decasia 's the visual film , the film in might be puzzle dull . | **CatGAN** | decasia is well below expectations and neither too erotic nor very effective . |

## 5. Conclusions

This paper proposes a sequence-to-sequence FA-GAN for category text generation. FA-GAN is feature-aware and category-aware, which can controllably generate category texts which are realistic, novel, and highly diversified.

The proposed FA-GAN consists of two neural networks, which are a sequence-to-sequence generator and a discriminator with an additional category classification head. The generator of FA-GAN consists of a feature encoder, a category encoder, a word encoder, and a relational-memory-core-based decoder with the Gumbel SoftMax function. The feature encoder of the generator captures the contexts of conditioned sentences so that the generator of FA-GAN is feature-aware. With the introduced feature encoder, FA-GAN has a sequence-to-sequence structure to address mode collapse and improves the sentence diversity. The category encoder is the key enabler for controllable category text generation. The relational-memory-core-based decoder can improve the expressiveness of the generator, and its Gumbel SoftMax activation function solves the text discreteness. For the discriminator of FA-GAN, except for the original authenticity classification head, a category classification head is supplemented to provide categorical information to guide the generator to synthesize correctly classified texts. To facilitate the training of FA-GAN, in addition to the Wasserstein adversarial loss, the multi-class classification loss is also included in the adversarial training of the generator and the discriminator.

On various text classification datasets, the proposed FA-GAN is compared with 10 text augmentation techniques on 6 datasets. Overall, the proposed FA-GAN has achieved the state-of-the-art classification accuracy on all the datasets. The superior BLEU scores and the highest negative log-likelihood values of generated sentences empirically validate that FA-GAN can generate realistic, fluent, and diversified category texts. A case study on the movie review dataset (MR-20) showcases that the generated sentences from FA-GAN are feature-aware and can match the required categories.

*Policy Gradient*. https://doi.org/10.48550/ARXIV.1609.05473

Zeng, H., Liu, J., Wang, M., & Wei, B. (2021). A sequence to sequence model for dialogue generation with gated mixture of topics. *Neurocomputing*, *437*, 282–288. https://doi.org/10.1016/j.neucom.2021.01.014

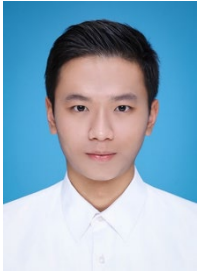

**Dr. Xinze Li** received his bachelor's degree in Electrical Engineering and its Automation from Shandong University, China, 2018. He received his Ph.D. degree in Electrical and Electronic Engineering from Nanyang Technological University, Singapore, 2023.

His research interests include dc-dc converter, modulation design, digital twins for power electronics systems, design process automation, light and explainable AI for power electronics with physics-informed systems, application of AI in power electronics, and deep learning and machine learning algorithms.

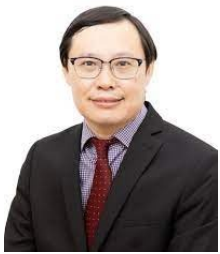

**Dr. Kezhi Mao** obtained his BEng, MEng and PhD from University of Jinan, Northeastern University, and University of Sheffield in 1989, 1992 and 1998 respectively. Since then, he has been working at School of Electrical and Electronic Engineering, Nanyang Technological University, Singapore, and is now an Associate Professor. His research covers a couple of subfields of artificial intelligence (AI), including machine learning, image processing, natural language processing and information fusion. Over the past 20 years, he has developed novel algorithms and frameworks to address various problems in AI. As a strong advocate of translational research, he has collaborated with government agencies and hospitals and developed a couple of prototypes of AI systems for image processing and natural language processing. He now serves as Member of Editorial Board of Neural Networks, Academic Editor of Computational Intelligence and Neuroscience, and General Chair and General Co-Chair of a number of international conferences.

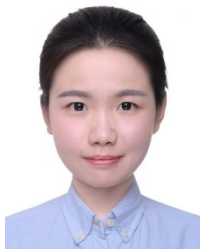

**Dr. Fanfan Lin** Fanfan Lin was born in Fujian, China in 1996. She received her bachelor degree in electrical engineering from Harbin Institute of Technology in China in 2018. She has been awarded the Joint Ph. D. degree in Nanyang Technological University, Singapore and Technical University of Denmark, Denmark, in 2023. Her research interest includes large language models for the design of large-scale power electronics systems, multi-modal AI for the maintenance of power converters, and the application of AI in power electronics.

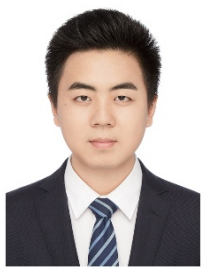

**Zijian Feng** received a B.Sc. degree from School of Electronics and Information Technology, Sun Yat-Sen University, Guangzhou, China, in 2019, and an M.Sc. degree in Signal Processing from School of Electrical and Electronic Engineering, Nanyang Technological University, Singapore, in 2020. He is currently pursuing a Ph.D. degree with the Interdisciplinary Graduate School, Nanyang Technological University, Singapore. His research interests include machine learning and natural language processing.